\documentclass{article}

\usepackage{arxiv}

\usepackage[utf8]{inputenc} 
\usepackage[T1]{fontenc}    
\usepackage{hyperref}       
\usepackage{url}            
\usepackage{booktabs}       
\usepackage{amsfonts}       
\usepackage{nicefrac}       
\usepackage{microtype}      
\usepackage{lipsum}
\usepackage{graphicx}
\usepackage{csquotes}
\usepackage{amsmath,amsfonts,amssymb}
\usepackage{isomath}
\usepackage{multirow}
\graphicspath{ {./images/} }

\title{A Comparison of Deep Saliency Map Generators on Multispectral Data in Object Detection}

\author{
 Jens Bayer \\
  Fraunhofer Center for Machine Learning\\
  Fraunhofer IOSB\\
  Gutleuthausstraße 1, 76275 Ettlingen, Germany \\
  \texttt{jens.bayer@iosb.fraunhofer.de} \\
   \And
 David Münch \\
  Fraunhofer Center for Machine Learning\\
  Fraunhofer IOSB\\
  Gutleuthausstraße 1, 76275 Ettlingen, Germany \\
  \And
 Michael Arens \\
  Fraunhofer Center for Machine Learning\\
  Fraunhofer IOSB\\
  Gutleuthausstraße 1, 76275 Ettlingen, Germany \\
}

\begin{document}
\maketitle
\begin{abstract}
Deep neural networks, especially convolutional deep neural networks, are state-of-the-art methods to classify, segment or even generate images, movies, or sounds. However, these methods lack of a good semantic understanding of what happens internally. The question, why a COVID-19 detector has classified a stack of lung-ct images as positive, is sometimes more interesting than the overall specificity and sensitivity. Especially when human domain expert knowledge disagrees with the given output. This way, human domain experts could also be advised to reconsider their choice, regarding the information pointed out by the system. In addition, the deep learning model can be controlled, and a present dataset bias can be found.

Currently, most explainable AI methods in the computer vision domain are purely used on image classification, where the images are ordinary images in the visible spectrum. As a result, there is no comparison on how the methods behave with multimodal image data, as well as most methods have not been investigated on how they behave when used for object 
detection. 

This work tries to close the gaps. Firstly, investigating three saliency map generator methods on how their maps differ across the different spectra. This is achieved via accurate and systematic training. Secondly, we examine how they behave when used for object detection.

As a practical problem, we chose object detection in the infrared and visual spectrum for autonomous driving. The dataset used in this work is the Multispectral Object Detection Dataset \cite{Takumi2017}, where each scene is available in the long-wave (FIR), mid-wave (MIR) and short-wave (NIR) infrared as well as the visual (RGB) spectrum. 

The results show that there are differences between the infrared and visual activation maps. Further, an advanced
training with both, the infrared and visual data not only improves the network's output, it also leads to more
focused spots in the saliency maps.
\end{abstract}


\section{Introduction}
\label{sec:intro}
Computer vision is no longer imaginable without convolutional deep neural networks. These methods provide
excellent performance in image classification, segmentation and object detection, and are best known 
in the non-academic area for their usage in autonomous driving, intelligent video surveillance 
and human medicine. The increasing integration into daily life led to a rising demand for 
explainability and interpretability of those black box models.
In recent years, a vast number of different methods \cite{Smilkov2017, Shrikumar2017, Sundararajan2017, Petsiuk2019, Wang2020a, Selvaraju2020, Fu2020, Muddamsetty2021, Bohle2021}
have been developed, that enable deep learning-based image classifiers
to provide a visual explanation of their classification result. 

Two big subsets of these explanation methods are gradient- and perturbation-based methods. 
The three investigated methods in this paper are also gradient- and perturbation-based:
Grad-CAM \cite{Selvaraju2020} is a method, that combines gradient information with the feature maps of the last convolution 
layer of a CNN, to generate a saliency map for a specific output class. RISE \cite{Petsiuk2019} is a Monte Carlo method, that
perturbs the input image and queries the network with the masked versions of the image. The resulting saliency map is 
the combination of each mask weighted by the corresponding prediction probability of the targeted class.
Like Grad-CAM, the third method SIDU \cite{Muddamsetty2021} uses the feature maps of the last convolution layer. These feature 
maps serve as masks for the input data. Similar to RISE, SIDU uses the network's output to calculate a similarity difference 
and uniqueness score for each masked input. The final saliency map is the weighted sum of the feature maps, where the weights are 
given by the corresponding similarity differences and the uniqueness scores.

\begin{figure}
    \centering
    \includegraphics[width=0.3\textwidth]{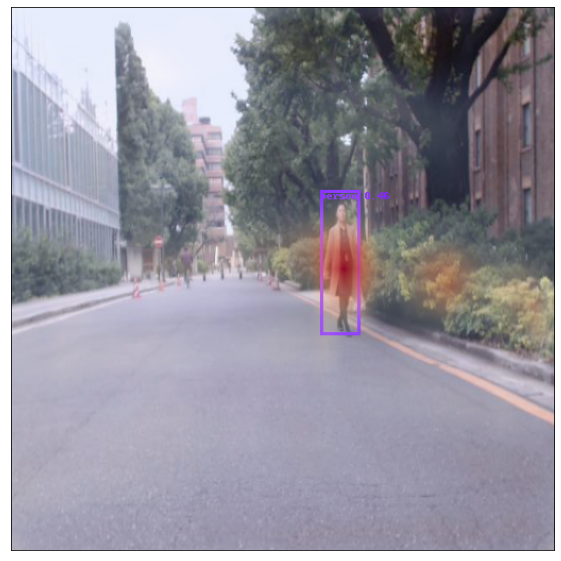}
    \includegraphics[width=0.3\textwidth]{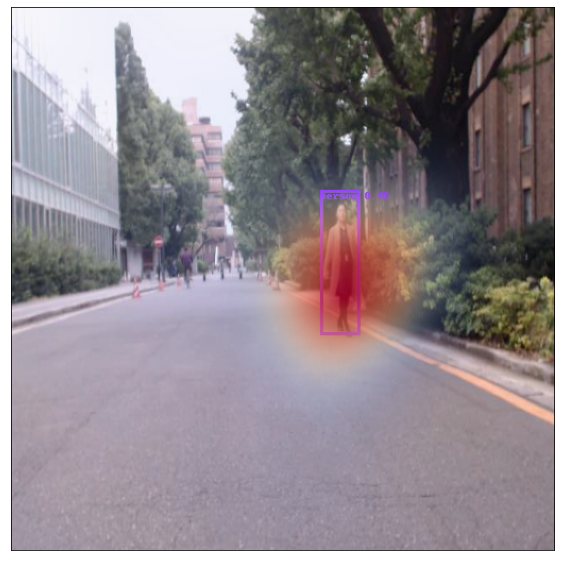}
    \includegraphics[width=0.3\textwidth]{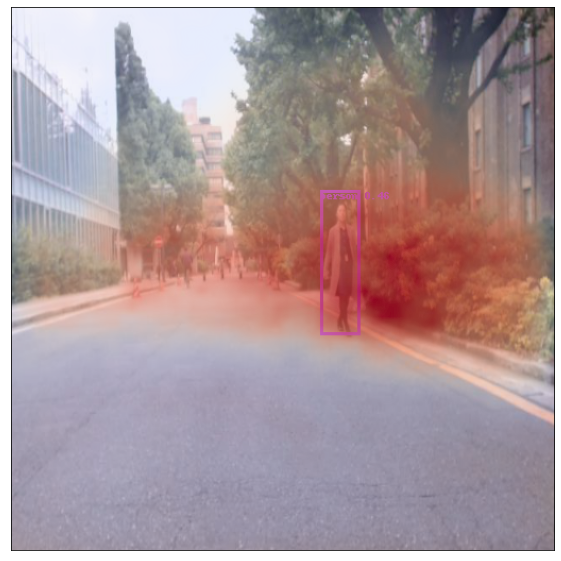}
    \caption{Saliency maps for the given bounding box, generated by Grad-CAM, RISE and SIDU.}
    \label{fig:tv_examples}
\end{figure}

The three methods are commonly used for the classification task. We investigate not only how these methods perform in the 
different spectra, but we also use them to generate saliency maps for the predicted bounding boxes (see \autoref{fig:tv_examples}).
To the best of our knowledge, this is also the first comparison of saliency map generators in object detection.

\section{Related Work}
\label{sec:relatedwork}
The following section covers selected work on object detection and explainable AI methods in the
deep learning domain. For a comprehensive overview of shallow models, the reader is referred to a survey paper for object detection
\cite{Shantaiya2013,Zou2019} or explainable AI methods \cite{Dosilovic2018, Burkart2020, Tjoa2020}.
\subsection{Object Detection}
Deep learning-based object detection can generally be divided into one-stage and two-stage detectors \cite{Zou2019,Liu2020a}. One-stage
detectors predict bounding boxes directly from the given input image. Two-stage detectors have a preprocessing step to generate classification and regression proposals \cite{Jiao2019}.

Faster R-CNN\cite{Ren2017} is a two-stage object detector and introduced the Region Proposal Network (RPN). 
RPNs are fully convolutional networks, that predict bounding boxes as well as objectness scores.
An RPN uses the attention \cite{Chorowski2015} mechanism and is responsible for the higher throughput in comparison to 
the earlier Fast R-CNN \cite{Girshick2015} approach.
The Single Shot MultiBox Detector (SSD) \cite{Liu2016} is a one-stage object detector, that \enquote{discretizes the 
output space of bounding boxes into a set of default boxes over different aspect ratios and scales per 
feature map location}. 
YOLOv4 \cite{Bochkovskiy2020} is the fourth version of the famous one-stage detector framework. 
It combines a variety of \enquote{Bag-of-Freebies} and \enquote{Bag-of-Specials} like 
CIoU Loss \cite{Zheng2020} or Mish activation \cite{Misra2019} to further improve speed and accuracy.
With EfficientNet\cite{Tan2019} as a backbone, and their proposed bi-directional feature pyramid network, 
EfficientDet \cite{Tan2020} is a one-stage detector that is scalable and performant while maintaining high accuracy.

Recently, the research focus has shifted to the utilization of the underlying mechanisms of the transformer architecture
\cite{Carion2020,Zhu2020,Perreault2020}.

\subsection{Explainable AI}
Most deep learning methods suffer from missing interpretability and explainability. Despite their potential good performance on test and
validation data, the poor understanding of the internal processes leads to a lack of confidence in these methods. For the 2D image
classification, there are methods to highlight areas in the image that are of special interest for the decision-making process of a
convolutional neural network. 
These methods can generally be categorized into gradient-based, perturbation-based and other underlying mechanics.

Class Activation Maps (CAM)\cite{Zhou2016} are one of the earlier attempts to highlight the activations of the network. CAM requires 
a network to use global average pooling just before the fully connected layers. The final saliency map is then given by the weighted 
sum of the feature maps of the final convolution layer, where the weights are extracted from the fully connected layer for the 
target class.

Gradient-weighted Class Activation Mapping (Grad-CAM) \cite{Selvaraju2020} is a generalization of CAM and does not require a global
average pooling. Instead, the method weights the upsampled feature maps of the ultimate convolution layer according to the 
gradient.

Sundararajan et al. \cite{Sundararajan2017} identify two axioms --- Sensitivity and Implementation Invariance --- that attribution methods 
should satisfy and present Integrated Gradients, a simple to implement method, that requires no modification to 
the original network and fulfills the named axioms. Integrated Gradients aggregates gradients along the straight line 
path between a reference image and the input image.

A combination of gradient- and perturbation-based method is SmoothGrad \cite{Smilkov2017}. 
SmoothGrad samples Gaussian noised variations of the input image and averages the resulting gradient of these. 
In comparison to the vanilla gradient, the averaged gradient is much smoother
and less noisy.

Randomized Input Sampling for Explanation (RISE) \cite{Petsiuk2018} is a perturbation-based method. The input image is masked with 
several random masks and propagated through the network. The final saliency map is then obtained as the weighted sum of the random masks,
where the weights are given according to the output of the network for the corresponding perturbed input.

A similar yet different approach is SIDU \cite{Muddamsetty2021}. During the forward propagation, the feature maps of the last convolution
layer are extracted. The input is then masked with the extracted, upsampled feature maps and propagated through the network. For the
resulting network outputs, a so-called similarity difference and uniqueness score\cite{Muddamsetty2021} is calculated. The upsampled 
feature maps are then weighted according to the similarity difference and uniqueness scores and summed. The result is the saliency map
for the predicted output.

Since DeconvNets \cite{Zeiler2014} and Guided-Backpropagation \cite{Springenberg2015} have been proven \cite{Nie2018} to just do partial image reconstruction, they are not described further.

\section{Saliency Map Generators}
\label{sec:saliency}
Saliency Map Generators visualize the activations of a convolutional neural
network. Given an input image $\tensorsym{I}$ and a saliency map generator $f(\tensorsym{X})$, the
saliency map $\matrixsym{S} = f(\tensorsym{I})$ highlights the most crucial areas for the decision process.
In the following Section, we give a brief introduction about the examined saliency map generators:
Grad-CAM \cite{Selvaraju2020}, SIDU \cite{Muddamsetty2021} and RISE \cite{Petsiuk2019}. 

\subsection{Grad-CAM}
\label{sec:gradcam}
Convolutional layers retain spatial information across consecutive layers. As a result, the ultimate convolution layer 
in a feed forward network contains the most dense spatial information of the image \cite{Selvaraju2020}. Based on the output feature maps of 
this layer, the input images are usually classified.

Grad-CAM uses the output features of this last convolution layer, to generate a saliency map.
The method is a generalization of Class Activation Mapping (CAM) \cite{Zhou2016} and requires neither a fully convolutional
network, nor a global average pooling layer \cite{Selvaraju2020}. Instead, a weighted sum of the extracted feature maps 
from the last convolution layer is calculated and bilinear upsampled.

Let $\tensorsym{F} \in \mathbb{R}^{N \times w \times h}$ be the output feature map of the ultimate convolution layer $l$. 
We then propagate the gradient $\tensorsym{G}^c$ for the target output class $c$ back to $l$. Finally, the saliency map 
\begin{equation}
    \matrixsym{S}^c = \sum_{i=1}^{N} ReLU(\alpha^c_i \cdot \matrixsym{F}_i)
    \label{eq:gradcam}
\end{equation} can be calculated, where the weights
\begin{equation}
    \alpha^c_i = \frac{1}{w \cdot h} \sum_{u}^{w} \sum_{v}^{h} \tensorsym{G}^c_{u,v}
\end{equation}
for each feature map $\matrixsym{F}_i$ are given as the mean of the corresponding gradients.

\subsection{RISE}
Instead of internal feature maps or gradient information, RISE perturbs the input and uses 
the output class probabilities of the network for the perturbed input as weights for the 
resulting saliency map. Compared to Grad-CAM, RISE has a significantly higher computational cost. 
The reason for this is that RISE requires a forward propagation for each perturbation mask $\matrixsym{M}_i$.
First, $N$ small binary masks are sampled according to a distribution $\mathcal{D}$. 
The masks are then bilinear upsampled to be slightly larger, then the input size. Afterwards, 
the masks are randomly cropped to match the input size. For each generated 
mask $\matrixsym{M}_i$, the network output $\vectorsym{p}_i$ of the masked input image 
\begin{equation}
\label{eq:maskedinput}
    \tensorsym{\tilde{I}}_i = \matrixsym{M}_i \circ \tensorsym{I}
\end{equation}
is calculated. The saliency map 

\begin{equation}
    \matrixsym{S}^c = \frac{1}{\mathbb{E}[\tensorsym{M}] \cdot N} \sum_{i=1}^{N} \vectorsym{p}^c_i \cdot \matrixsym{M}_i
\end{equation}
is then given as the sum of the upsampled masks, weighted by the corresponding probability score of the selected class and normalized 
by the expectation of M.

\subsection{SIDU}
In contrast to Grad-CAM and RISE, SIDU is incapable of generating a saliency map for a specific class. Instead, the
generated saliency map highlights the image areas with the highest impact on the output feature vector. To achieve this,
SIDU uses the feature maps $\matrixsym{F}_i$ of the last convolution layer to mask the input data. Those feature
maps are bilinear upsampled to match the input size. Afterwards, the Hadamard product of 
the input $\tensorsym{I}$ and each mask $\matrixsym{M}_i$
is propagated through the network. The saliency map is the weighted sum of the masks, where 
the weights are calculated according to the masks impact on changes in the similarity difference 
\begin{equation}
    sd_i = exp(\frac{-1}{2 \sigma^2} \cdot \lVert \vectorsym{p}_o - \vectorsym{p}_i \rVert)
\end{equation}
and uniqueness scores
\begin{equation}
u_i = \sum_{j=1}^{N} \lVert \vectorsym{p}_i - \vectorsym{p}_j \rVert
\end{equation}
of the network's output $\vectorsym{p}$, regarding the unmasked input:
\begin{equation}
    S^c = \sum_{i=1}^{N}  sd_i \cdot u_i \cdot \matrixsym{M}_i
\end{equation} 
Here, $\vectorsym{p}_o$ is the network's output for the unmasked input.

\section{Experimental Setup}
\label{sec:experiments}

\subsection{EfficientDet}
For each spectrum, an EfficientDet-D0\footnote{\url{https://github.com/zylo117/Yet-Another-EfficientDet-Pytorch.git}} 
is trained. The networks are trained
via Adam \cite{Kingma2015} over thirty epochs, with an initial learning rate of 0.001 and a 
learning rate reduction by a factor of ten each tenth epoch. The used batch size is eight.
The raw images of the dataset are split with 
a ratio of 1:4 in a test and train set and resized to a fixed size of $512\times512$
pixel. The single channel of the infrared images is tripled to be compatible with the 
network's input layer.

For the evaluation, the trained detectors are applied for each input image of the 
corresponding test set. For each image, the bounding box with the highest score 
is selected and investigated. Since SIDU and RISE sample the network multiple 
times with modified input data, the selected bounding box has to be found again. 
This is achieved via a search over all bounding boxes with a score of at least 
0.05 and an intersection over union (IoU) of at least 0.5. The bounding box with the highest IoU is selected as the corresponding bounding box.

For the calculation of the saliency maps, the network's output, especially the bounding box
scores, are used. The feature maps that are required by Grad-CAM and SIDU are extracted directly from 
the classifier head. Since the ReLU activation in \autoref{eq:gradcam} leads to empty or almost empty
activation maps, \autoref{sec:results} contains also the results without the ReLU activation.
RISE uses 500 masks with a sample size of $8\times8$ and a probability of $0.1$ for a pixel to be non-zero.

\subsection{Dataset}
\begin{figure}[htb]
    \centering
    \includegraphics[width=0.45\textwidth]{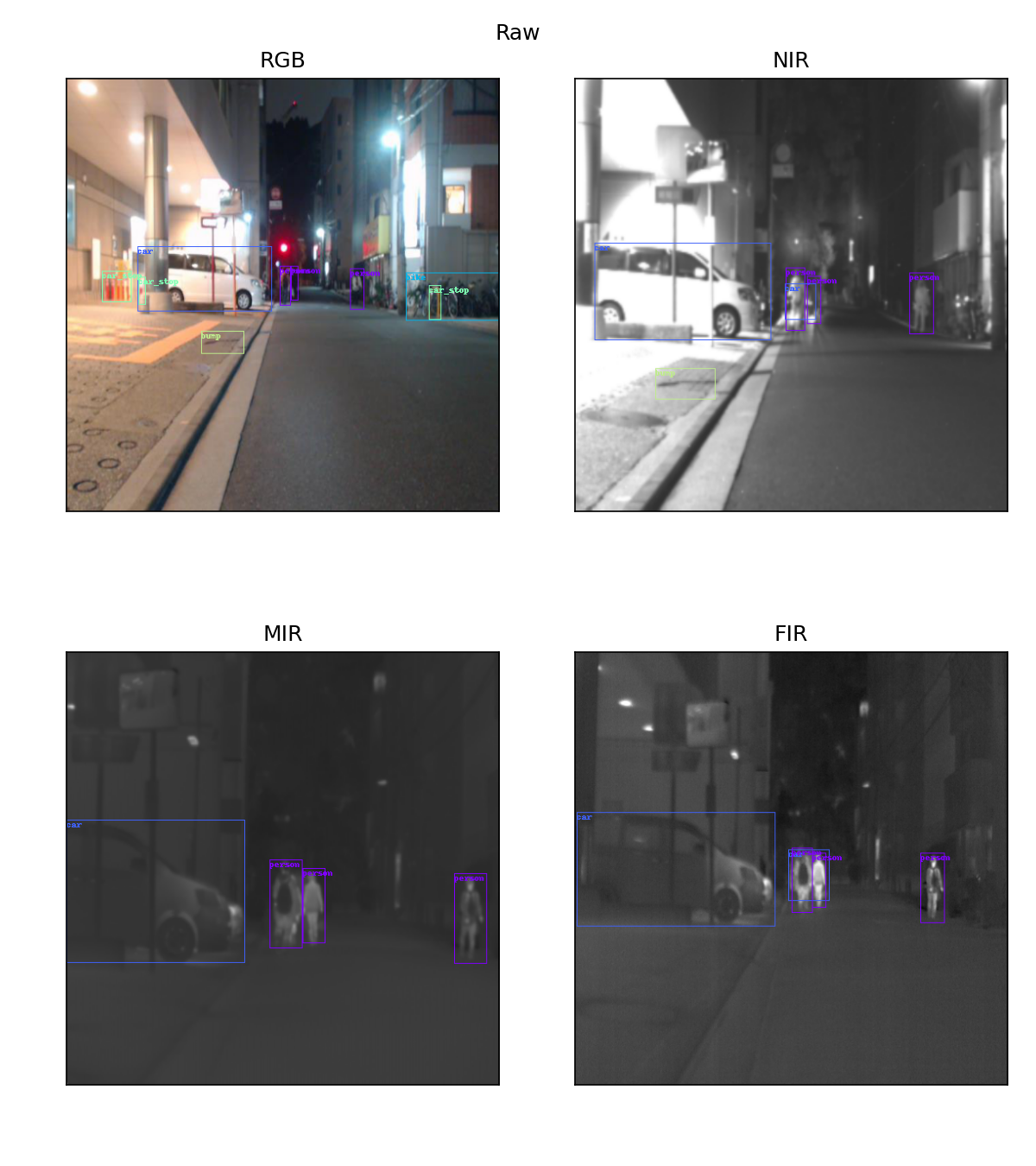}
    \includegraphics[width=0.45\textwidth]{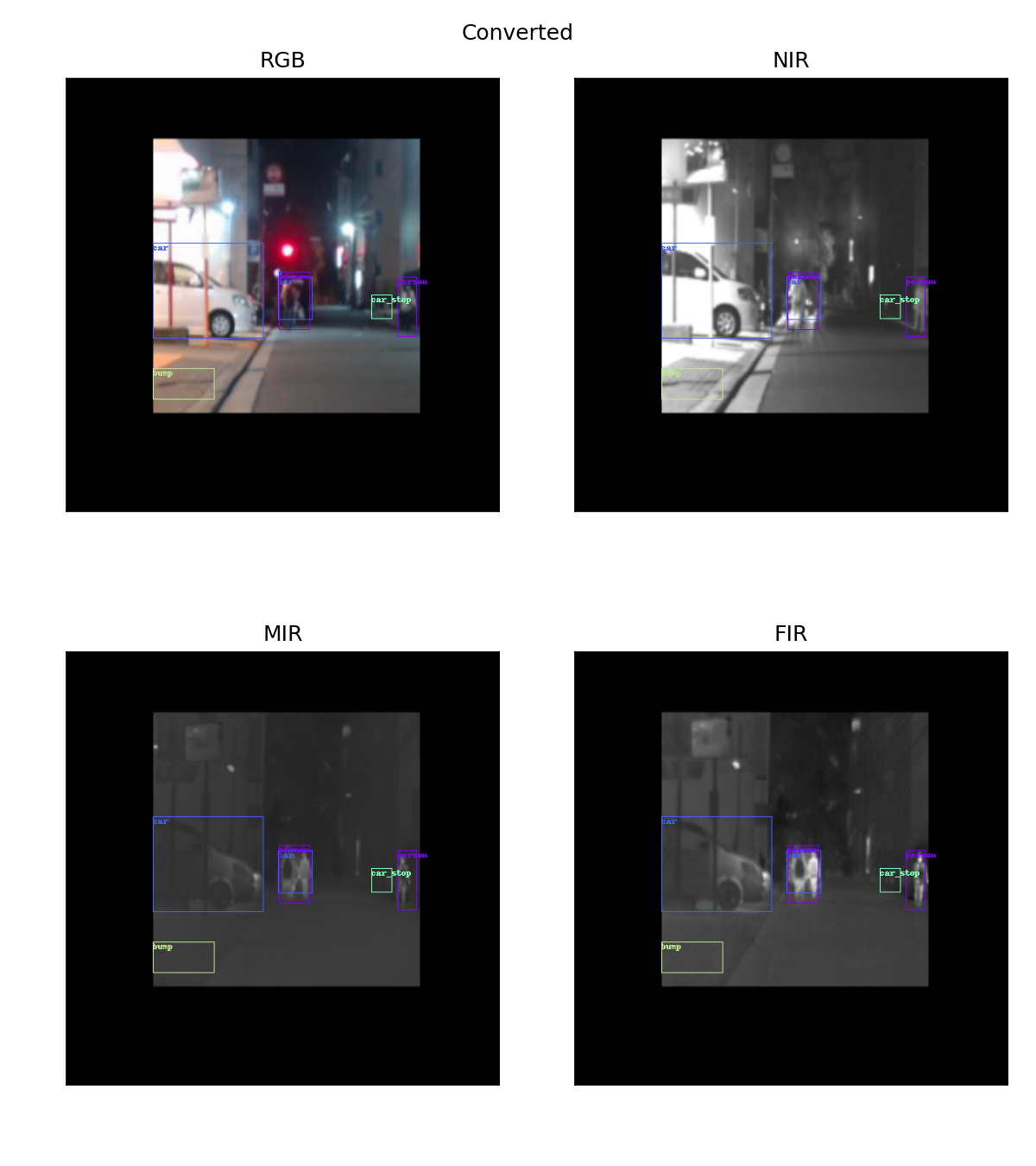}
    \caption{Examples of the Multispectral Dataset \cite{Takumi2017}. The same scene in different spectra: 
    RGB, NIR, MIR and FIR. The raw images, which are used for the experiments, are given on the left half. 
    Converted images with a common viewpoint are given on the right half.}
    \label{fig:dataset}
\end{figure}

To ensure an accurate and systematic training, we used the Multispectral Object Detection Dataset \cite{Takumi2017}. 
The dataset contains sequences, recorded in a university environment during day and nighttime. The sequences were 
captured with four cameras (RGB, NIR, MIR, FIR) and a frame rate of one frame per second. Since the
viewpoints of the cameras are different, the dataset also contains converted versions of the images, where they 
share a common viewpoint (see \autoref{fig:dataset}).

\autoref{tbl:image_sizes} contains the resolution of the images for the different spectra. 
\begin{table}[htb]
    \centering
    \begin{tabular}{|c|c|c|}
         \hline
         Spectrum & Width (px) & Height (px)\\
         \hline
         RGB & 640 & 480 \\
         NIR & 320 & 256 \\
         MIR & 320 & 256 \\
         FIR & 640 & 480 \\
         \hline
    \end{tabular}
    \caption{Raw image sizes of the Multispectral Object Detection dataset.}
    \label{tbl:image_sizes}
\end{table}
The ground truth is given in form of bounding boxes with the corresponding classes: 
Person, car, bike, color\_cone, car\_stop, bump, hole, animal, unknown.

\section{Evaluation}
\label{sec:evaluation}
For a quantitative comparison of the methods, the insertion, and deletion metrics\cite{Petsiuk2019} are used.
The deletion metric removes pixels of the input successively according to their importance for a given
saliency map. As a result, the probability of the predicted class decreases, the more pixels are removed.
If the threshold is plotted against the output probability, the area under the resulting curve gives
information about the performance of the method. If the most crucial pixels for the decision are removed 
first, there is a sharp drop in the curve and the area under the curve is close to zero.
For the insertion metric, clear pixels are successively added to a blurred version of the original input image.
In the optimal case, the area under the curve of the resulting plot is close to one and has 
therefore an early sharp rise.
\autoref{sec:results} contains plots of the deletion and insertion metrics for the different methods and
spectra.

As stated in \autoref{tab:results}, the lowest total deletion score is achieved with SIDU and 
NIR data ($0.10\pm0.03$). The highest total insertion score is also achieved with SIDU, but with 
RGB+MIR data ($0.67\pm0.21$). The results also show that Grad-CAM has a significant higher mean 
deletion score and standard deviation than the other two methods. This is similar to the lower
mean insertion score of Grad-CAM. Both, SIDU and RISE achieve similar results for the deletion
and insertion metric, whereas SIDU performs slightly better.

\begin{figure}
    \centering
    \includegraphics[width=\textwidth]{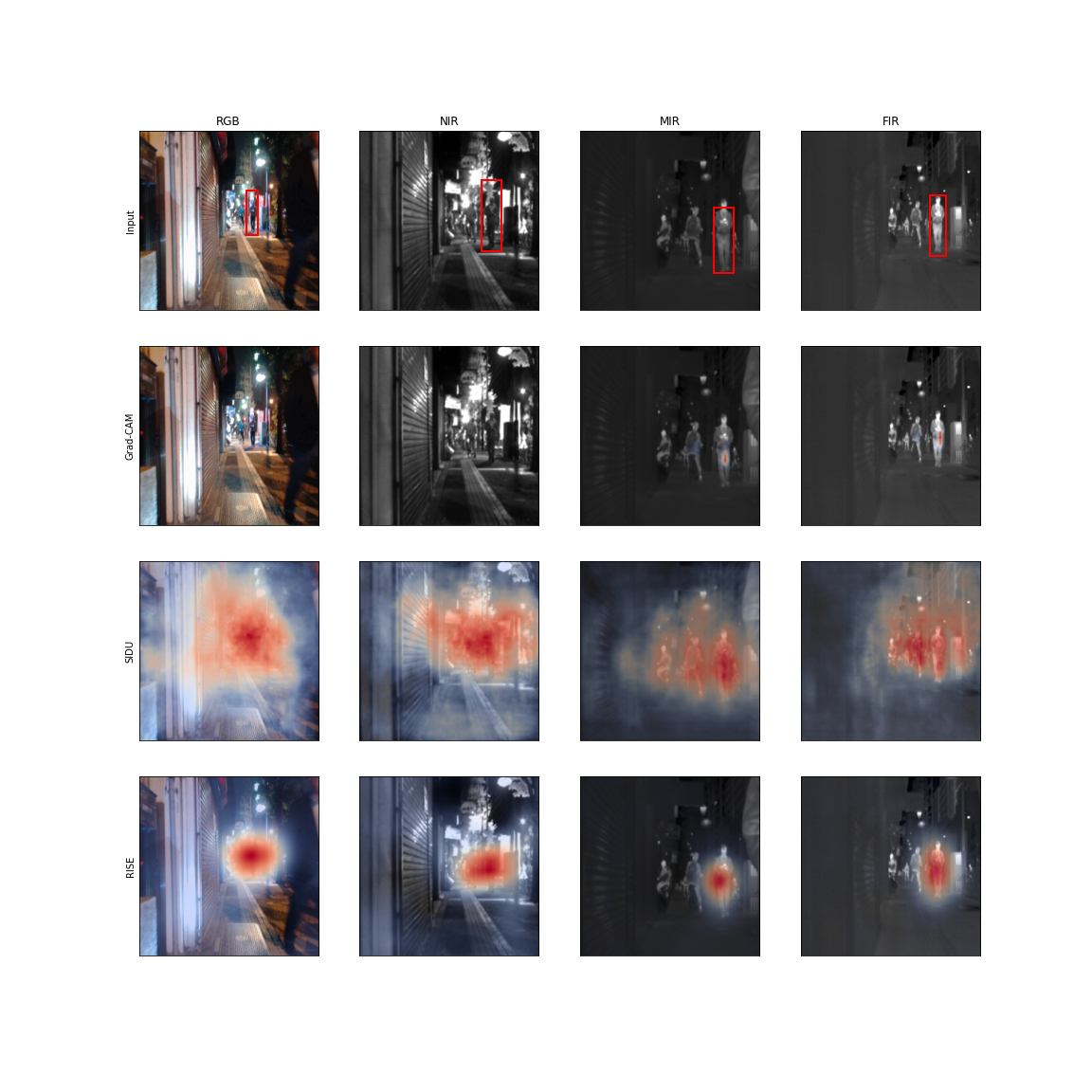}
    \caption{Generated saliency maps for raw images of different spectra. The top row shows the original images with the 
    queried bounding box. The following rows show the saliency maps, generated by Grad-CAM, RISE and SIDU.}
    \label{fig:example_results}
\end{figure}

\autoref{fig:example_results} shows generated saliency maps for a bounding box in the same scene but different spectra. 
Qualitatively, the saliency maps of Grad-CAM are mostly a single sharp focus spot. The reason
for this is the application of the ReLU in \autoref{eq:gradcam}. \autoref{fig:tv_examples} and the
figures in \autoref{sec:results} are examples of Grad-CAM saliency maps, without the ReLU activation.
A larger, more diffuse single focus spot is given in the saliency maps, generated by RISE.
Like Grad-CAM, the queried bounding box always contains the center of the focus spot. 
Saliency maps generated via Grad-CAM without ReLU have multiple 
smaller focus spots, where the main focus is almost always in the queried bounding box as well. 
SIDU, on the other hand, generates much diffuser saliency maps, that are significant larger.

\begin{figure}
    \centering
    \includegraphics[width=\textwidth]{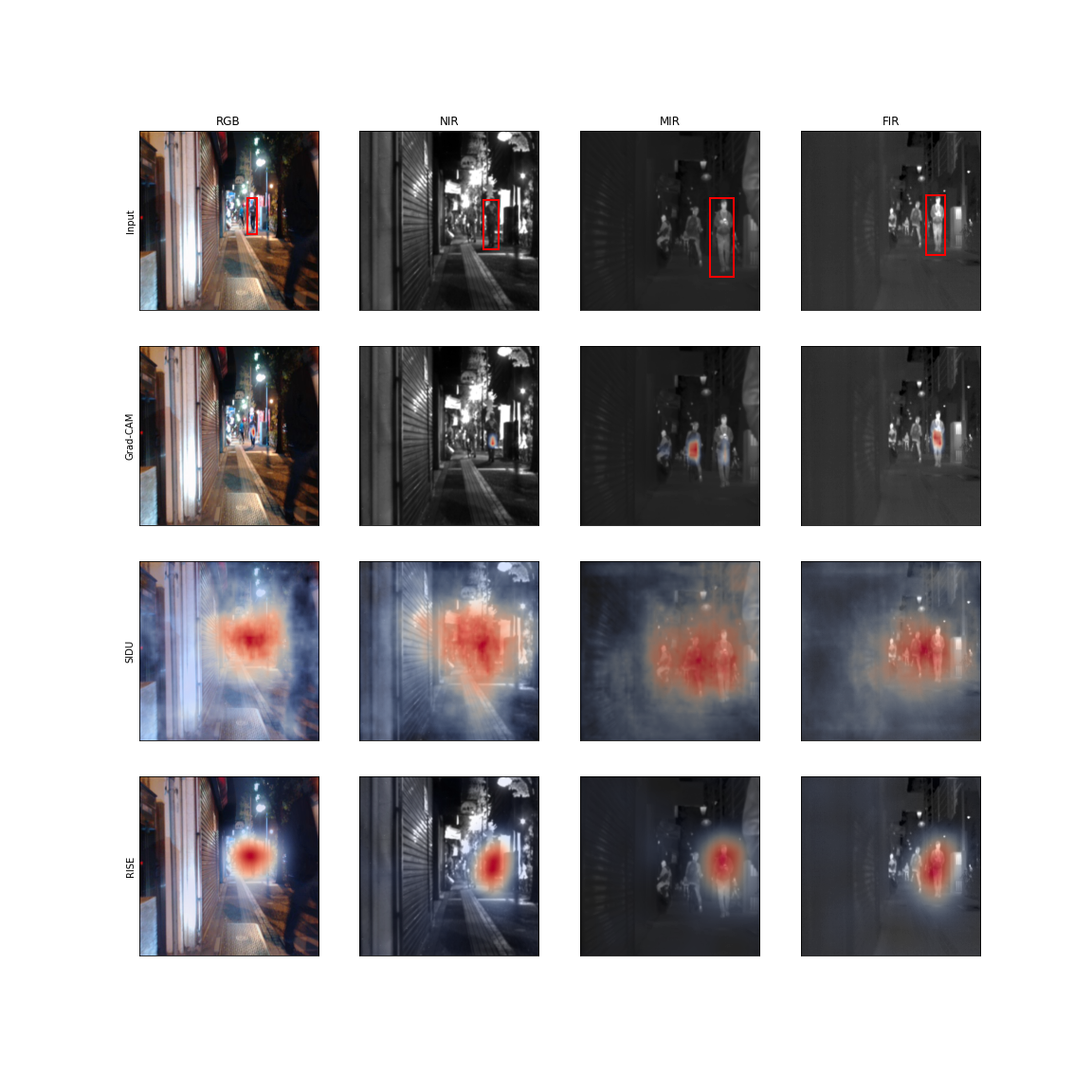}
    \caption{Generated saliency maps for raw images of different spectra with the trained RGB+MIR network. 
    The top row shows the original images with the queried bounding box. 
    The following rows show the saliency maps, generated by Grad-CAM, RISE and SIDU.}
    \label{fig:example_results_rgb+mir}
\end{figure}

\autoref{fig:example_results_rgb+mir} shows the saliency maps for the same scene as in 
\autoref{fig:example_results}, but this time generated by a single network trained with 
RGB and MIR data. For Grad-CAM, the saliency maps have become larger. Especially, the maps 
for RGB and NIR are now visible. 
The saliency maps generated by RISE have become smaller and adapt more to the shape of the
object inside the bounding box.
SIDU also profits from the extended training: The saliency maps are a bit smaller and contain more focused areas.
Nonetheless, the saliency maps of SIDU still seem to focus random areas. The reason for this is, that SIDU in its
original implementation is not using any additional information of the target like RISE or Grad-CAM. When using SIDU
in a classification task, the resulting saliency map can not be generated for a specified class. Instead, the method
\enquote{visual explanations for the prediction} \cite{Muddamsetty2021}.

When comparing RGB and infrared saliency maps, the focus areas of the RGB saliency maps are diffuser than the infrared ones.

In terms of computation speed, Grad-CAM (0.1 seconds) is much faster than SIDU (8.5 seconds) or RISE (32 seconds, 5000 masks, 24 mask batch)
on our test system (Intel i9-9980XE, Nvidia Quadro RTX 6000). For a real-time application, Grad-CAM is therefore the only applicable approach.


\begin{table}
    \centering
    \begin{tabular}{|c|c|c|}
        \hline
         & Deletion $\downarrow$ & Insertion $\uparrow$ \\
         \hline
         Grad-CAM RGB & $0.22\pm0.24$ & $0.30\pm0.14$\\
         RISE RGB     & $0.13\pm0.05$ & $0.34\pm0.13$\\
         SIDU RGB     & $0.11\pm0.04$ & $0.38\pm0.14$\\
         \hline
         \hline
         Grad-CAM NIR & $0.23\pm0.24$ & $0.34\pm0.14$\\
         RISE NIR     & $0.12\pm0.05$ & $0.34\pm0.13$\\
         SIDU NIR     & $\mathbf{0.10\pm0.03}$ & $0.38\pm0.13$\\
         \hline
         \hline
         Grad-CAM MIR & $0.47\pm0.26$ & $0.47\pm0.21$\\
         RISE MIR     & $0.15\pm0.06$ & $0.54\pm0.20$\\
         SIDU MIR     & $0.13\pm0.03$ & $0.58\pm0.18$\\
         \hline
         \hline
         Grad-CAM FIR & $0.53\pm0.29$ & $0.52\pm0.22$\\
         RISE FIR     & $0.17\pm0.07$ & $0.61\pm0.20$\\
         SIDU FIR     & $0.15\pm0.05$ & $0.61\pm0.20$\\
         \hline
         \hline
         Grad-CAM RGB + MIR & $0.60\pm0.31$ & $0.53\pm0.27$\\
         RISE RGB + MIR     & $0.25\pm0.12$ & $0.64\pm0.22$\\
         SIDU RGB + MIR     & $0.20\pm0.08$ & $\mathbf{0.67\pm0.21}$\\
         \hline
    \end{tabular}
    \caption{Mean and standard deviation of the deletion and insertion metrics for the different methods and models on 
    evaluation data. The $\uparrow$/$\downarrow$ arrow indicates, whether a higher or lower score is desired.}
    \label{tab:results}
\end{table}

\section{Conclusion}
\label{sec:conclusion}
This paper compares deep saliency map generators in object detection when used with multispectral data. 
As an object detector, EfficientDet has exemplarily been investigated. The experiments and evaluation show that the 
methods can easily be applied to the given task and result in similar but slightly distinguishable saliency maps
for the different spectra. Grad-CAM and RISE show, most of the time, a single clear focus on the area around the
questioned bounding box, while SIDU seems to also highlight unnecessary areas.
When trained on both, RGB and MIR images, the main focus of the saliency methods is more focussed than
on the separated training. 
Although the quantitative evaluation shows, that SIDU achieves the lowest deletion and highest insertion
score, RISE offers the most obvious saliency map.
In terms of usability in a real-time environment, only Grad-CAM can be calculated within a reasonable  time.
Further research could investigate how these methods perform on volumetric or spatio-temporal 
input data like CT scans or videos.

\section*{Acknowledgments}
This work was developed in Fraunhofer Cluster of Excellence \enquote{Cognitive Internet Technologies}.

\bibliographystyle{unsrt}  
\bibliography{references}  


\appendix
\section{Results}
\label{sec:results}
\begin{figure}[!htb]
    \centering
    \includegraphics[width=0.3\textwidth]{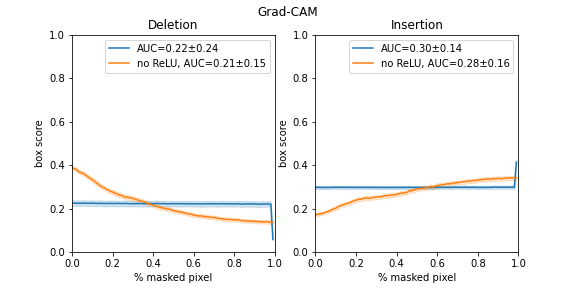}
    \includegraphics[width=0.3\textwidth]{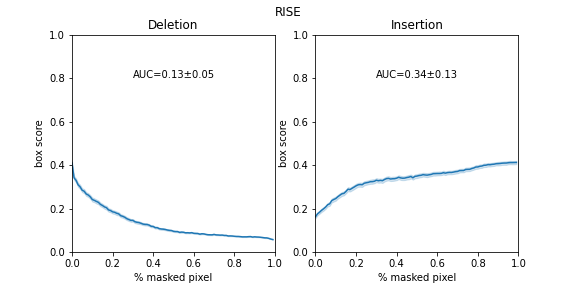}
    \includegraphics[width=0.3\textwidth]{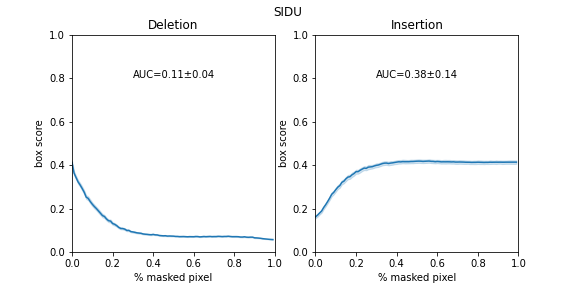}
    \caption{Insertion and Deletion evaluation of Grad-CAM, RISE and SIDU on RGB data.}
    \label{fig:eval_rgb}
\end{figure}
\begin{figure}[!htb]
    \centering
    \includegraphics[width=0.3\textwidth]{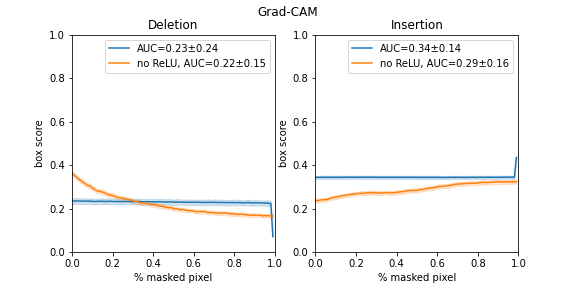}
    \includegraphics[width=0.3\textwidth]{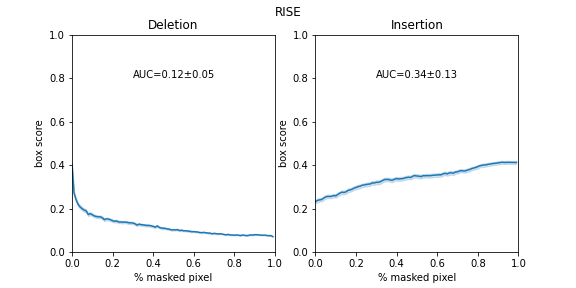}
    \includegraphics[width=0.3\textwidth]{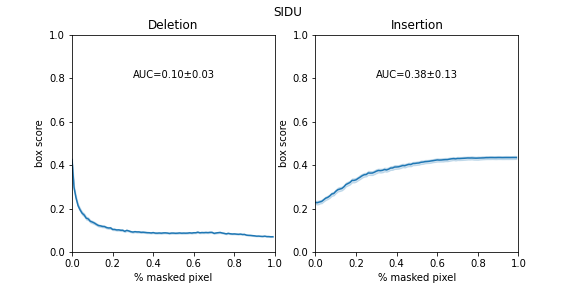}
    \caption{Insertion and Deletion evaluation of Grad-CAM, RISE and SIDU on NIR data.}
    \label{fig:eval_nir}
\end{figure}

\begin{figure}[!htb]
    \centering
    \includegraphics[width=0.3\textwidth]{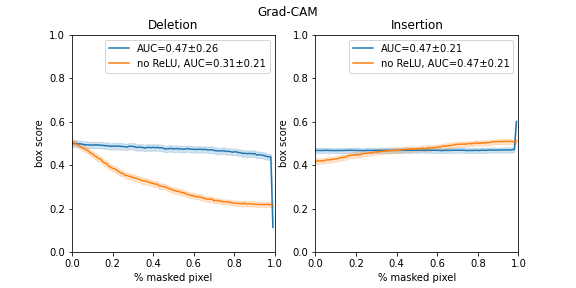}
    \includegraphics[width=0.3\textwidth]{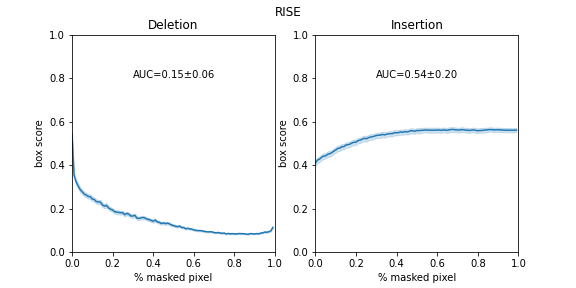}
    \includegraphics[width=0.3\textwidth]{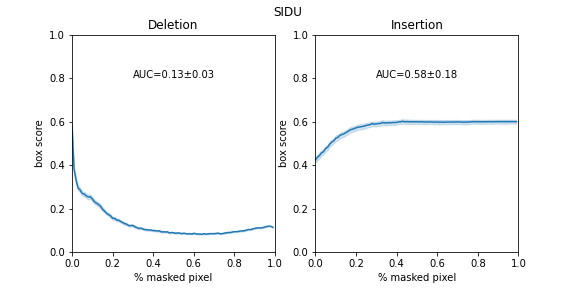}
    \caption{Insertion and Deletion evaluation of Grad-CAM, RISE and SIDU on MIR data.}
    \label{fig:eval_mir}
\end{figure}
\begin{figure}[!htb]
    \centering
    \includegraphics[width=0.3\textwidth]{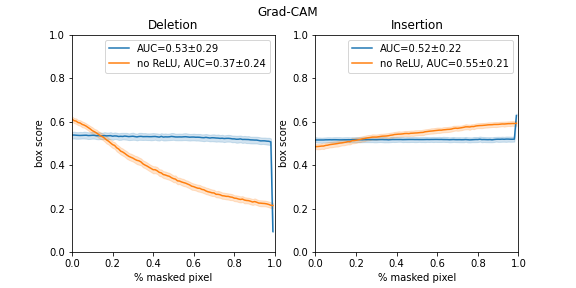}
    \includegraphics[width=0.3\textwidth]{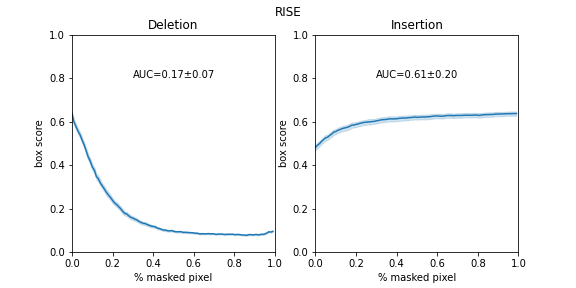}
    \includegraphics[width=0.3\textwidth]{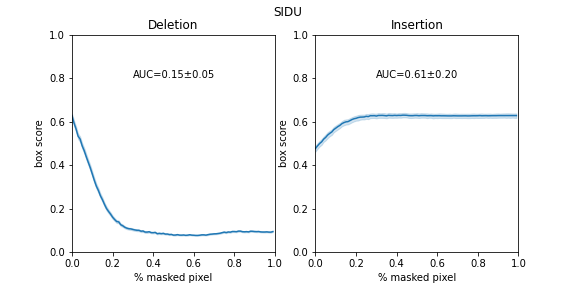}
    \caption{Insertion and Deletion evaluation of Grad-CAM, RISE and SIDU on FIR data.}
    \label{fig:eval_fir}
\end{figure}
\begin{figure}[!htb]
    \centering
    \includegraphics[width=0.3\textwidth]{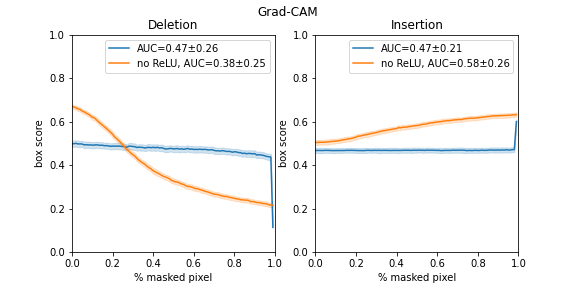}
    \includegraphics[width=0.3\textwidth]{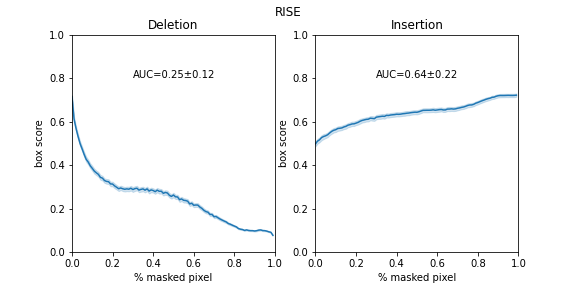}
    \includegraphics[width=0.3\textwidth]{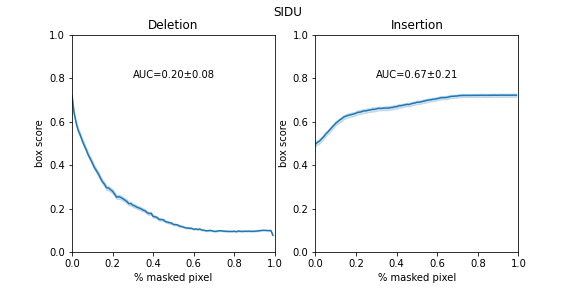}
    \caption{Insertion and Deletion evaluation of Grad-CAM, RISE and SIDU on RGB + MIR data.}
    \label{fig:eval_RGB_MIR}
\end{figure}

\begin{figure}
    \centering
    \includegraphics[width=\textwidth]{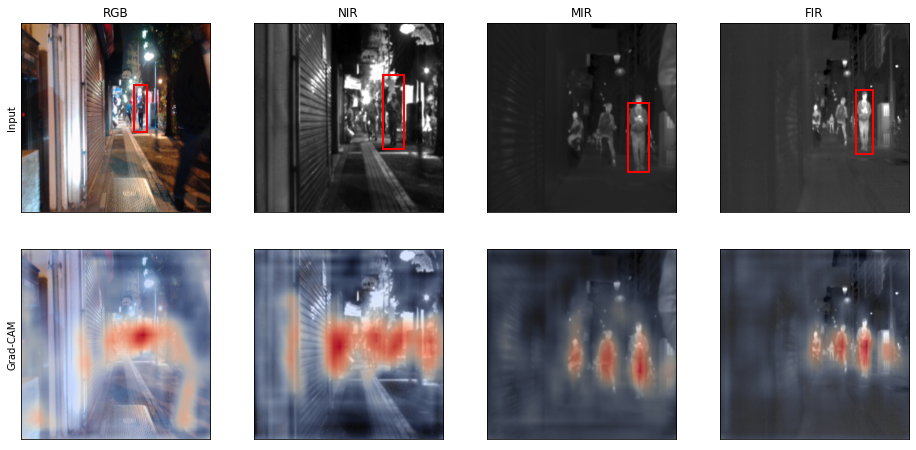}
    \caption{Generated saliency maps for raw images of different spectra. The top row shows the original images with the 
    queried bounding box. The second row shows the saliency maps, generated by Grad-CAM without the ReLU activation.}
    \label{fig:example_results_norelu}
\end{figure}
\begin{figure}
    \centering
    \includegraphics[width=\textwidth]{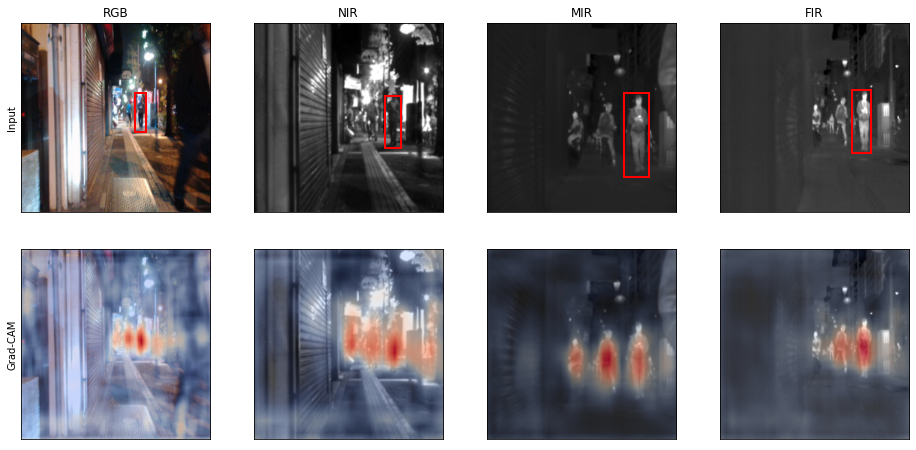}
    \caption{Generated saliency maps for raw images of different spectra with the trained RGB+MIR network. 
    The top row shows the original images with the queried bounding box. 
    The second row shows the saliency maps, generated by Grad-CAM without the ReLU activation.}
    \label{fig:example_results_rgb+mir_norelu}
\end{figure}

\end{document}